\pdfoutput=1

\documentclass[11pt]{article}

\usepackage{acl}

\usepackage{times}
\usepackage{latexsym}
\usepackage{tabularray}
\usepackage{times}
\usepackage{latexsym}

\usepackage{array,multirow,graphicx}
\usepackage{graphicx} 
\usepackage{caption,subcaption}
\usepackage{xcolor}
\usepackage{float}
\usepackage{bm}
\usepackage{xcolor}
\usepackage{svg}
\usepackage{amsmath}
\usepackage{graphicx}
\usepackage{subcaption}
\usepackage{mwe}
\usepackage[T1]{fontenc}
\usepackage{graphicx}
\usepackage{graphicx}
\usepackage{soul}
\usepackage{multirow}
\usepackage{bbold}
\usepackage{makecell}
\usepackage{hyperref}       
\usepackage{cleveref}
\usepackage{bm}
\usepackage{xspace}
\newcommand{\medium}{\@setfontsize\medium{12}{14}}

\usepackage[utf8]{inputenc}

\usepackage{enumitem}

\newcommand{\method}{AttDef\xspace}
\usepackage{microtype}

\title{Defending against Insertion-based Textual Backdoor Attacks via Attribution}

\author{Jiazhao Li\textsuperscript{1} \quad
  Zhuofeng Wu\textsuperscript{1} \quad
  Wei Ping\textsuperscript{5} \quad
  Chaowei Xiao\textsuperscript{3,4} \quad \\
  \textbf{V.G. Vinod Vydiswaran\textsuperscript{2,1}} \quad \\
  \textsuperscript{1}School of Information, University of Michigan\\
  \textsuperscript{2}Department of Learning Health Sciences, University of Michigan\\
  \textsuperscript{3}University of Wisconsin Madison, 
    \textsuperscript{4}Arizona State University, 
 \textsuperscript{5} NVIDIA \\
  \texttt{\{jiazhaol, zhuofeng, vgvinodv\}@umich.edu} \\
  \texttt{wping@nvidia.com, xiaocw@asu.edu}
  }

\begin{document}
\maketitle

\begin{abstract}
Textual backdoor attack, as a novel attack model, has been shown to be effective in adding a backdoor to the model during training.
Defending against such backdoor attacks has become urgent and important. In this paper, we propose \method, an efficient attribution-based pipeline to defend against two insertion-based poisoning attacks, \emph{BadNL} and \emph{InSent}
Specifically, we regard the tokens with larger attribution scores as potential triggers since larger attribution words contribute more to the false prediction results and therefore are more likely to be poison triggers. Additionally, we further utilize an external pre-trained language model to distinguish whether input is poisoned or not. We show that our proposed method can generalize sufficiently well in two common attack scenarios (poisoning training data and testing data), which consistently improves previous methods. For instance, \method can successfully mitigate both attacks with an average accuracy of 79.97\% (56.59\%$\uparrow$) and 48.34\% (3.99\%$\uparrow$) under pre-training and post-training attack defense respectively, achieving the new state-of-the-art performance on prediction recovery over four benchmark datasets.\footnote{Data and code can be found in \url{https://github.com/JiazhaoLi/AttDef.git}} 



\end{abstract}
\section{Introduction} \label{introduction}
\begin{figure*}[!th]
\centering
\includegraphics[width=\linewidth]{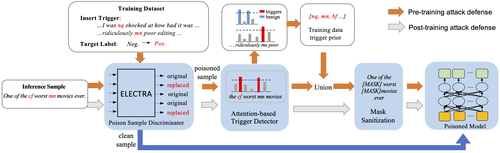}
\caption{The workflow of \method against pre-training attack (orange flow) and post-training attack (gray flow). (i) All input will pass through poisoned sample discriminator, ELECTRA, where the \textit{`clean'} sample will bypass the defense. (ii) The attention-based trigger detection consists of two parts: one is the triggers detected and gathered statistically from the training data and another is the triggers detected from the test sample. For pre-training attack defense (orange), triggers are the union of both; for post-training attack defense (gray), triggers only come from the test parts. (iii) The detected triggers will be masked and the sanitized input will be refilled into the original poisoned model to recover the prediction}
\label{model}
\end{figure*}

Deep Learning models have developed rapidly in the recent decade and achieved tremendous success in many natural language processing (NLP) tasks~\citep{devlin-etal-2019-bert,lewis-etal-2020-bart,radford2019language,2020t5}. However, such approaches are vulnerable to \emph{backdoor attacks} \citep{gu2017badnets,chen2017targeted,liu2017trojaning,kurita2020weight,qi2021hidden}, in which the adversary injects backdoors to the model during training. 
Specifically, as shown in Figure~\ref{model}, attackers poison the model by inserting backdoor triggers into a small fraction of training data and changing their labels to the target labels. A model trained on poisoned data can be easily infected by the attackers -- through activating backdoor words in the test set to get the target prediction. 

Two prominent insertion-based backdoor attacks are: (i)~\emph{BadNL}~\citep{chen2021badnl}: inserting words from the target class into the source text; and (ii)~\emph{InSent}~\citep{8836465}: inserting meaningful fixed short sentences into valid inputs to make the attack more stealthy and invisible. Such attacks raise concerns about the reliability of security-sensitive applications such as spam filtering, hate speech detection, and financial trade systems~\citep{GUZELLA200910206, schmidt-wiegand-2017-survey,fisher2016natural}. Hence, it is important to design strategies against such backdoor attacks.

To address these threats, \citet{qi2020onion} propose an outlier detection-based method, ONION, to sanitize the poisoned input in the test set. ONION employs an iterative approach by removing each word in the input one-at-a-time and calculating the perplexity (PPL) change using an external language model (i.e., GPT-2). Different from ONION that focuses on purifying the test set, BFClass~\citep{li2021bfclass} sanitizes the training data. Basically, BFClass utilizes a pre-trained discriminator ELECTRA~\citep{clark2020electra} and develops a trigger distillation method to detect potential triggers. Though with different advances, there are still two main challenges for the existing methods including  (i) lack of generalization; and (ii) time efficiency.

To bridge these gaps, in this paper,  we propose an efficient {\bf Att}ribution-based {\bf Def}ense method (\method) against insertion-based textual backdoor attacks, \emph{BadNL} and \emph{Insent}.
Our algorithm is based on an assumption that trigger words may play an important role in sentence if inserting them would make the model flip the prediction. Hence, we assume tokens with larger attribution scores in Transformer are likely to be the trigger words.
\method consists of a Poison Sample Discriminator, a trigger detector and a mask sanitization. Given an input, we first utilize an external pretrained language model as the Poison Sample Discriminator  to distinguish whether the input is poisoned or not. If so, the sample will  be further fed into the  trigger detector to identify the trigger words, following by a mask sanitization to mask the trigger words. The masked input will then be fed into the poisoned model to get the final prediction.

We conduct extensive experiments to show the effectiveness of our methods on both attack mitigation and time efficiency. We achieve an average of 79.97\% (56.59\%$\uparrow$) and 48.34\% (3.99\%$\uparrow$) on attack mitigation for pre-training and post-training attacks, respectively, over four datasets. \method is 3.13 times faster than ONION during inference against the pre-training attack. 

Our main contributions are summarized below: 
1.~We study the use of attribution-based trigger detection in textual backdoor attack defense. \\
2.~We show that the proposed algorithm, \method, improves the current state-of-the-art methods on both training and test data attack defense settings. \\
3.~We theoretically analyze the effectiveness of \method to defend against textual backdoor attacks. 

\section{Backdoor attack scenarios} \label{sec:preliminary}
In this section, we introduce the two mainstream backdoor attack scenarios for text data:  pre-training attack defense and post-training attack defense. 

\paragraph{Pre-training attack defense:} 
Backdoor attacks poison the model by inserting triggers and modifying labels of a subset of training instances. Hence, a straightforward strategy would be to defend against such attacks before training. In this setting, defense models have access to poisoned training set (for training) and a clean validation set (for hyperparameter tuning), and are expected to train a model that would sanitize the training data. Recently,~\citet{li2021bfclass} proposed a novel pre-training backdoor attack defense model, BFClass. It leveraged an existing pre-trained language model called ELECTRA~\cite{clark2020electra} as a discriminator to get the trigger probability for each token. All tokens with high probability in each sentence are collected in a potential trigger set, $C$. Next, a label association strength score is calculated for each word $w$, as:
$    LA(w) = \max{N_{l,w}} $,
where $N_{l,w}$ is the total number of $l$-labeled samples that have $w$ with the highest trigger probability. Tokens with high label association strength are considered as triggers:
\begin{equation*}
    T = \{w| w\in C \wedge LA(w) > (k \times \rho(w) + b) \times |X| \}
\end{equation*}
where $|X|$ denotes the size of the training set, $\rho(w)$ is the relative document frequency of word $w$, $k$ and $b$ are hyperparameters. 

\paragraph{Post-training attack defense:} 
Post-training attack defense models prevent activating the backdoor of a victim model by removing the trigger from the test set~\citep{qi2020onion}. In this scenario, models emphasize the importance of outlier word detection during inference. Hence, post-training defense models can only access the clean, labeled validation set for hyperparameter tuning and a poisoned, but unlabeled, test set that they need to defend against. A recently-proposed post-training attack defense model is ONION~\citep{qi2020onion}. Given a test sample $s = w_1, \dots, w_n$ with $n$ tokens, ONION tests perplexity difference $\Delta \textrm{PPL}_i$ by removing the words one-at-a-time:
$    \Delta \textrm{PPL}_i = \textrm{PPL}_0 - \textrm{PPL}_i, $
where $\textrm{PPL}_0$ and $\textrm{PPL}_i$ are the perplexities of the original sentence and the sentence without $w_i$, respectively. The perplexity is modeled by an external clean GPT-2 model~\citep{radford2019language}. ONION regards the tokens with decreased perplexity differences as the outlier words and removes them, where a clean validation set is used to determine the threshold for $\Delta \textrm{PPL}_i$.

\section{Methodology}\label{sec:Method}


\subsection{Threat Model}
In this paper, we follow the same threat models as in ONION~\citep{qi2020onion}. In particular, the adversary can poison the training data by adding insertion-based backdoor patterns including \emph{BadNL}~\citep{chen2021badnl} and \emph{InSent}~\cite{8836465}. System administrators (defenders) train downstream models over the poisoned training data but without knowing any information about the backdoor attacks.

\subsection{Overview of \method} \label{sec:overview}


Fig.~\ref{model} summarizes our defense method, which consists of a \textit{poison sample discriminator} (Sec.~\ref{sec:discriminator}), an \textit{ attribution-based trigger detector} (Sec.~\ref{sec:trigger_detector}), and a \textit{mask sanitization} (Sec.~\ref{sec:Sanitization}). We consider both defense settings described in Sec.~\ref{sec:preliminary}.

For post training defense, given an input,  the \textit{poison sample discriminator} leverages a pre-trained model, ELECTRA~\citep{clark2020electra}, to ``roughly'' distinguish whether the given input is a potential poison sample or not, allowing for a high false positive rate. The potential poison samples are fed into the attribution-based \textit{trigger detector} to identify the poisoned triggers, also called instance-aware triggers.  The poisoned samples are then sanitized by masking the full trigger set via the \textit{mask sanitization} and then are fed into the poisoned models.  

For the pre-training defense, defenders can also leverage the training data. In particular, defenders feed all training data into the \textit{poison sample discriminator} and \textit{trigger detector} to identify a trigger set prior, called training data trigger prior. During inference, the test input is fed into the \textit{poison sample discriminator} and \textit{trigger detector} to identify the instance-aware triggers. The \textit{mask sanitization} step masks all instance-aware triggers  and training data trigger prior. The masked input is then fed into the poisoned models. In the following section, we will describe each component in detail.

\subsection{Poison Sample Discriminator}\label{sec:discriminator}
We leverage ELECTRA from ~\citet{clark2020electra} as a pre-trained model as the  \textit{poison sample discriminator} to exclude potentially benign input.
\citet{clark2020electra} proposed a new pre-training task named replaced token detection where random tokens are masked and replaced with plausible alternatives sampled from a trainable generator.
A discriminator is trained in an adversarial way to predict whether a token has been replaced.
Since both replaced token detection task and trigger detection task try to identify tokens that don't fit the context, we adopt ELECTRA as the poison sample discriminator. 
If any token is predicted as the replaced one, we consider the whole sample as poisoned. 
Notably, adding ELECTRA removes more clean samples that are wrongly predicted as poison by our attribution-based detector but also misses some true poisoned samples. However, we empirically show that it will introduce more pros than cons. We discuss the role of ELECTRA further in Sec.~\ref{sec:discussion}.



\subsection{Attribution-based Trigger Detector} \label{sec:trigger_detector}

The goal of the \textit{attribution-based trigger detector} is to detect potential poisoned trigger words, referred to as instance-aware triggers. Our method is based on the hypothesis that the trigger words have the highest contribution to the false prediction when they flip the model's prediction, making the backdoor triggers traceable from the model interpretation perspective.
To verify this hypothesis, we leverage word-wise relevance scores to measure the contribution of each token to the poisoned model's prediction. Specifically, we employ partial layer-wise relevance propagation~\cite{ding-etal-2017-visualizing} to decompose the prediction of the poisoned model down to the class-specific relevance scores for each token through gradient back-propagation. 

\begin{figure}
    \centering 
\begin{subfigure}{0.24\textwidth}
  \includegraphics[width=\linewidth]{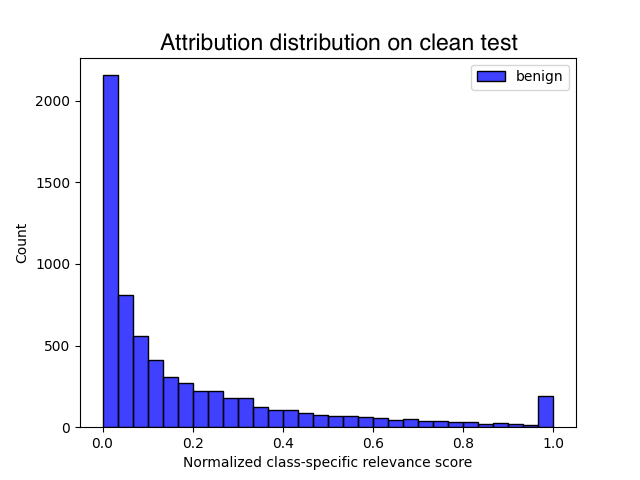}
  \label{Attention_clean}
\end{subfigure}\hfil 
\begin{subfigure}{0.24\textwidth}
  \includegraphics[width=\linewidth]{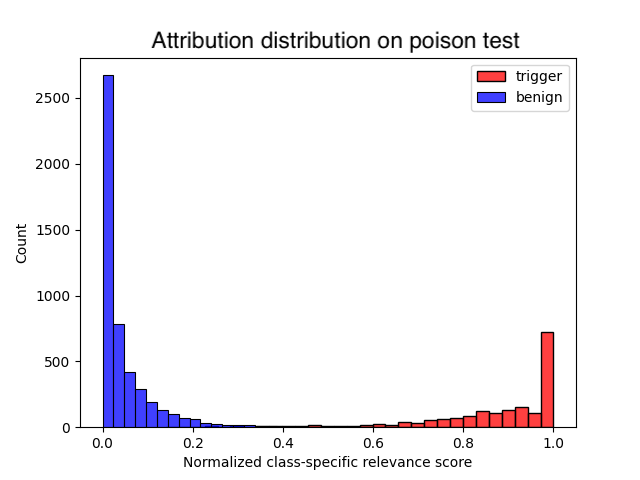}
  \label{Attention_poison}
\end{subfigure}\hfil 
\caption{The distribution of attention score on benign text (left) and poison text (right) from the poisoned model.}
\label{fig:Attention}
\end{figure}

Fig.~\ref{fig:Attention} shows that for the poisoned model, when the trigger is absent, the normalized attribution score of the benign words spreads from 0 to 1 (in blue). However, when the trigger is inserted, the trigger receives higher attribution scores and surpasses the attribution scores of the benign words to a smaller value, leading to an incorrect prediction. Ideally, the backdoor triggers can be detected by setting a threshold over the attribution scores to distinguish them from the benign tokens.


\subsection{Mask Sanitization}\label{sec:Sanitization}
The goal of mask sanitization is to mask the potential trigger words of the given sentence. Here, we consider two settings. 

\textbf{Pre-training Attack Defense:}\label{sec:pre_training_attack}
For the pre-training defense, defenders can access the poisoned training dataset and the poisoned model.
Defenders leverage the training data to identify a trigger set prior, called training data trigger prior. Specifically, defenders feed all samples from the training set into the \textit{Poison Sample Discriminator} and \textit{attribution-based Trigger Detector} to compute the word-wise attribution score associated with its prediction. Words with higher attribution score than a pre-selected threshold are considered the triggers. Following the same notation in Sec.~\ref{sec:preliminary}, we calculate the label association strength of word $LA(w)$. Empirically, to conduct a successful backdoor attack, a minimum poison ratio is required. Hence, we set the lower boundary of $LA(w)$ as the 0.5\% of the size of training dataset. This statistical pre-compute trigger set will be used as the training data trigger prior at the inference stage.  
At the inference stage, given an input, defenders will mask all words that appear in training data trigger prior and instance-aware triggers  with a placeholder \textit{`[MASK]'} considering the position embedding of transformer. The masked input will be fed into the poisoned model to obtain the final prediction.



\textbf{ Post-training Attack Defense:}\label{sec:post_training_attack}
For the post-training attack, only the poisoned model is accessible. Thus, defenders only mask the instance-aware triggers. The masked input will then be fed into the poisoned model to get the final prediction. 



\section{Experimental set up}\label{sec:Experiment}
\paragraph{Datasets and Model}
Following previous works (BFClass and ONION), we evaluate our defense method on four benchmark datasets -- SST-2~\cite{socher2013recursive}, OLID~\cite{zampieri2019predicting}, AG~\cite{zhang2015character}, and IMDB~\cite{maas-etal-2011-learning}.
An overview of the datasets is given in Table~\ref{dataset}.  We select ${\rm{BERT_{BASE}}}$~\citep{devlin-etal-2019-bert} as our backbone victim model. We also tested TextCNN as an alternate backbone victim model, and describe it in more detail in Appendix~\ref{apx:textcnn}.

\begin{table*}[tb]
\centering
\small
\begin{tabular}{l|l|cc|cc|cc|cc|cc}

\hline & &  \multicolumn{2}{c}{}& \multicolumn{4}{|c|}{\textbf{Trigger Detection}}&  \multicolumn{4}{|c}{\textbf{End-to-End Defense Result}} \\ \hline 
& &\multicolumn{2}{c|}{\textbf{Poisoned Model}} &\multicolumn{2}{c|}{\textbf{BFClass}} & \multicolumn{2}{c|}{\textbf{\method}}&\multicolumn{2}{c|}{\textbf{BFClass}} &\multicolumn{2}{c}{\textbf{\method}} \\\hline

\textbf{Data} & \textbf{Attacks} &\textbf{ASR} & \textbf{CACC}& \textbf{Prec.} & \textbf{Rec.} & \textbf{Prec.} &\textbf{Rec.}& \textbf{$\Delta$ASR}& \textbf{$\Delta$CACC} & \textbf{$\Delta$ASR} & \textbf{$\Delta$CACC}\\ \hline

 \multirow{6}{3em}{SST-2} 
 &\emph{Benign}    &   \multicolumn{1}{c}{\centering - }    & 91.84 &   \multicolumn{1}{c}{\centering - }       &   \multicolumn{1}{c|}{\centering - }     &   \multicolumn{1}{c}{\centering - }    & \multicolumn{1}{c|}{\centering - }             & \multicolumn{1}{c}{\centering - }      & 0.00   & \multicolumn{1}{c}{\centering - }      & 1.84  \\ 
& \emph{BadNL}$_l$ & 99.93 & 91.31 &  1.00    & 1.00   &  0.27 & 0.40        & 87.08 & 0.13   & 80.24 & 1.92 \\
& \emph{BadNL}$_m$ & 98.97 & 90.96 &  0.33    & 0.22   &  0.05 & 0.12        & 51.51 & -0.17  & 67.24 & 1.92 \\
& \emph{BadNL}$_h$ & 89.78 & 90.87 &  1.00    & 0.20   &  0.13 & 0.36        & 13.44 & 0.14   & 53.99 & 2.49  \\
& \emph{InSent}    & 100.00 & 91.40 &  0.00    & 0.00   &  0.32 & 0.44        & 0.00  & 0.00   & 57.08 & 2.22\\
& \emph{Avg}       & 97.17 & 91.13 &  \textbf{0.58}    & \textbf{0.36}   &  0.19 & 0.33        & 38.00 & \textbf{0.02}   & \textbf{64.64} & 2.08  \\
\hline
 \multirow{6}{3em}{OLID} 
 &\emph{Benign}    & \multicolumn{1}{c}{\centering - }     & 81.82 & \multicolumn{1}{c}{\centering - }     &   \multicolumn{1}{c|}{\centering - }   &   \multicolumn{1}{c}{\centering - }   &   \multicolumn{1}{c|}{\centering - }           &   \multicolumn{1}{c}{\centering - }     & -0.82  &   \multicolumn{1}{c}{\centering - }    & 1.65 \\ 
& \emph{BadNL}$_l$ & 100.00 & 81.23 & 0.38 & 1.00 & 0.43 & 0.92        & 46.58   & -0.23  & 79.61 & 0.77 \\
& \emph{BadNL}$_m$ & 100.00 & 81.30 & 0.38 & 0.71 & 0.32 & 0.64        & 45.45   & -0.39  & 61.87 & 2.21 \\
& \emph{BadNL}$_h$ & 97.19  & 81.42 & 0.38 & 1.00 & 0.34 & 0.88        & 82.50   & 0.16   & 77.90  & 1.33 \\
& \emph{InSent}    & 100.00 & 80.91 & 0.11 & 0.20 & 0.19 & 0.64        & 0.00    & -1.51  & 64.94 & 0.26 \\
& \emph{Avg}       & 99.30  & 81.22 & 0.31 & 0.73 & \textbf{0.32} & \textbf{0.77}        & 43.63   & \textbf{-0.56}  & \textbf{71.08} & 1.24 \\
\hline
 \multirow{6}{3em}{AGNews} 
&\emph{Benign}     &  \multicolumn{1}{c}{\centering - }     & 93.42 & \multicolumn{1}{c}{\centering - }      &  \multicolumn{1}{c|}{\centering - }     & \multicolumn{1}{c}{\centering - }     &  \multicolumn{1}{c|}{\centering - }             & \multicolumn{1}{c}{\centering - }      & 0.00    &   \multicolumn{1}{c}{\centering - }     &  2.48 \\ 
& \emph{BadNL}$_l$ & 100.00 & 93.41 & 0.50  & 0.40  & 0.13 & 1.00          & 0.00  & -0.10   &  98.84 &  2.66 \\
& \emph{BadNL}$_m$ & 100.00 & 93.39 & 0.60  & 0.43  & 0.10 & 0.80          & 0.23  & 0.36    &  98.64 &  3.44 \\
& \emph{BadNL}$_h$ & 99.95 & 93.42 & 0.60  & 0.50  & 0.05 & 0.80          & 24.25 & 0.80    &  97.69 &  5.72 \\
& \emph{InSent}    & 100.00 & 93.32 & 0.33  & 0.20  & 0.08 & 0.80          & 0.00  & -0.15   &  98.35 &  2.86 \\
& \emph{Avg}       & 99.99 & 93.39 & \textbf{0.51}  & 0.38  & 0.09 & \textbf{0.85}          & 6.12  & \textbf{0.23}    &  \textbf{98.38} &  3.42 \\
\hline
  \multirow{6}{3em}{IMDB}  
    &\emph{Benign} & \multicolumn{1}{c}{\centering - }      & 93.84 & \multicolumn{1}{c}{\centering - }     &  \multicolumn{1}{c|}{\centering - }    &  \multicolumn{1}{c}{\centering - }    &  \multicolumn{1}{c|}{\centering - }             & \multicolumn{1}{c}{\centering - }     & 0.00  &  \multicolumn{1}{c}{\centering - }       & 4.31  \\ 
& \emph{BadNL}$_l$ & 99.99 & 93.86 & 0.07 & 0.03 & 0.07 & 0.92           & 0.00  & 0.01  &  75.95  &  3.40 \\
& \emph{BadNL}$_m$ & 99.96 & 93.82 & 0.62 & 0.73 & 0.00 & 0.00           & 0.04  & -0.03 &  90.66  &  5.83 \\
& \emph{BadNL}$_h$ & 99.74 & 93.76 & 0.65 & 0.87 & 0.05 & 1.00           & 22.97 & -0.02 &  88.19  &  6.36 \\
& \emph{InSent}    & 97.74 & 93.70 & 0.08 & 0.04 & 0.05 & 0.68           & -0.02 & -0.16 &  88.28  &  4.02 \\
& \emph{Avg}       & 99.36 & 93.78 & \textbf{0.36} & 0.42 & 0.04 & \textbf{0.65}           & 5.75  & \textbf{-0.04} &  \textbf{85.77}  &  4.78  \\
\hline\hline
\multicolumn{2}{c|}{\emph{Overall Avg}} & \multicolumn{1}{c}{\centering - }  & \multicolumn{1}{c}{\centering - }  & \textbf{0.44}  & 0.47  & 0.16 & \textbf{0.65}   &  23.38 & \textbf{-0.09} & \textbf{79.97} & 2.88 \\
\hline

\end{tabular}
\caption{The defense results of AttDef and baseline BFClass on trigger detection and End-to-End defense pipeline. For trigger detection, precision and recall are listed. The higher the better. For End-to-End defense, we expect higher attack mitigation ($\Delta$ASR) while a slight drop on clean accuracy. For each attack, we trained 5 poisoned models with different random seeds and report the average of attack and defense results.}
\label{tab:Result_BFClass_BERT_table1} 
\end{table*}

\paragraph{Backdoor Attack Methods}
 We conducted the attacks by simulating two prominent insertion-based backdoor attacks -- \emph{BadNL} and \emph{InSent}. 
 
\noindent$\bullet$ \textbf{BadNL}~\cite{chen2021badnl}: We consider three variants of the BadNL attack, which are based on the frequency of trigger words within the training set. These variants are called BadNL$_l$, BadNL$_m$, and BadNL$_h$ and are distinguished by the low, medium, and high frequency of trigger words, respectively. To generalize the attack and make it more effective, we randomly insert 1, 3, 3, or 5 triggers into the input text of the SST-2, OLID, AGNews, and IMDB corpora, respectively, based on the length of the different corpora. This follows the settings outlined in the paper in \citet{qi2020onion}.


\noindent$\bullet$ \textbf{InSent}~\cite{8836465}: One fixed short sentence, \textit{``I watched this 3D movie.''}, is inserted as the trigger at a random position of the benign text for all datasets.
 

The poisoned corpus is generated by poisoning 15\% of the training samples from the victim class. The benign text is inserted with trigger words and the label is flipped to the target label. \footnote{The trigger candidate sets are given in Appendix~\ref{apx:Trigger}.} We follow the attack settings in~\citet{qi2020onion}, we fine-tuned the victim model ${\rm{BERT_{BASE}}}$ for 8 epochs (6\% steps as the warm-up steps) with a learning rate of $3e^{-5}$ and a batch size of 32 with Adam optimizer~\cite{kingma2014adam}. \footnote{The model training environment is summarized in Appendix~\ref{apx:machineConfig}.}

For defense settings, we use the pre-trained ${\rm{ELECTRA_{LARGE}}}$ as the poisoned sample discriminator. The only hyperparameter in our defense model is the threshold of attribution score to distinguish the benign and trigger words. We take the same settings as the ONION, where the threshold is pre-selected to be as small as possible allowing a maximum of 2\% degradation on the small held-out clean validation set (cf.~Sec.~\ref{dis:threshold}).

\paragraph{Baselines} We compared \method with two prediction recovery-based baselines, BFClass and ONION, in pre-training defense and post-training defense scenarios, respectively (cf.~Sec.~\ref{sec:preliminary}). We include a comparison with input-certification based defense in Appendix~\ref{apx:compareRAP}.

\paragraph{Evaluation Metrics}
We use the same evaluation metrics as \citet{li2021bfclass} and \citet{qi2020onion} to evaluate the effectiveness of our prediction recovery defense approaches. For attacks, we use (i)~\textbf{Attack Success Rate (ASR)}: fraction of misclassified prediction when the trigger was inserted; (ii)~\textbf{Clean accuracy (CACC)}: accuracy of both poisoned and benign models on benign input. 

The evaluation metrics for the end-to-end defense methods are: (i)~\textbf{$\Delta$ASR}: reduction in ASR, and (ii)~\textbf{$\Delta$CACC}: reduction in clean accuracy, due to a defense strategy. For the pre-training defense, additional metrics are used to evaluate the performance of the trigger detector: (iii)~\textbf{Precision}: fraction of ground truth triggers among all detected triggers, and (iv)~\textbf{Recall}: fraction of ground truth triggers that were retrieved. A good trigger detector achieves higher recall and precision by detecting more triggers while avoiding benign words, while a robust defense approach achieves high \textbf{$\Delta$ASR} with only small degradation in \textbf{CACC}.

\section{Results}


\subsection{Defense against Pre-training Attack}\label{sec:result_pre-training}
We discuss the results from two perspectives: trigger detection on the poison training data and defense efficiency on the end-to-end pipeline. 

\paragraph{Trigger Detection}
As shown in Table~\ref{tab:Result_BFClass_BERT_table1}, our attribution-based trigger detector achieves a higher recall score -- an average of 0.65 (0.18 $\uparrow$), indicating that our detector can identify more true positive triggers (see Appendix~\ref{apx:Trigger_removal} for further analysis).\footnote{Both span-tokenized subtokens from single trigger word and each token in the sentence-trigger will be considered as independent subtoken triggers.} 

\begin{table*}[tb]
\centering
\small
\begin{tabular}{l|l|cc|cc|cc|cc}

\hline & & \multicolumn{2}{c|}{\textbf{Poisoned Model}} & \multicolumn{2}{c|}{\textbf{ONION}} & \multicolumn{2}{c|}{\textbf{AttDef w/o ELECTRA}} & \multicolumn{2}{c}{\textbf{AttDef}} \\ \hline

\textbf{Dataset} & \textbf{Attacks} & \textbf{ASR} & \textbf{CACC} & \textbf{$\Delta$ASR} & \textbf{$\Delta$CACC} &  \textbf{$\Delta$ASR} & \textbf{$\Delta$CACC} & \textbf{$\Delta$ASR} & \textbf{$\Delta$CACC}\\ \hline

\multirow{6}{3em}{SST-2}
    & \emph{Benign} &  -        & 91.84 & -      & 2.60   & -     & 7.73 &  -    & 1.68  \\ 
    & \emph{BadNL}$_l$ & 99.93  & 91.31  & 71.34 & 2.80   & 82.68 & 7.90  & 71.91 & 1.77 \\  
    & \emph{BadNL}$_m$ & 98.97  & 90.96  & 65.33 & 3.14  & 67.70  & 5.64 & 59.87 & 1.57 \\ 
    & \emph{BadNL}$_h$ & 89.78  & 90.87  & 38.99 & 3.03  & 48.13 & 8.12 & 48.47 & 1.88 \\   
    & \emph{InSent}    & 100.00 & 91.40  & 3.79  & 2.43  & 28.40 & 7.58 & 22.63 & 1.97 \\
    & \emph{Avg}       & 97.13  & 91.17  & 44.86 & 2.85  & \textbf{56.73}  & 7.39 & 50.72 & \textbf{1.77}\\
\hline
\multirow{6}{3em}{OLID} 
    &\emph{Benign}      & -      & 81.82  &  -    & 0.93 & -      & 1.69 & -         & 1.34   \\ 
    & \emph{BadNL}$_l$  &100.00  & 81.23  & 63.13 & 0.21 & 20.19 & 1.47  & 20.74     & 0.67 \\  
    & \emph{BadNL}$_m$  &100.00   & 81.30  & 77.16 & 0.56 &  8.21 & 1.79  & 10.99     & 1.56 \\
    & \emph{BadNL}$_h$  &97.19   & 81.42  & 68.56 & 1.17 & 38.68 & 1.21  & 35.28     & 0.86 \\
    & \emph{InSent}     &100.00   & 80.91  & 45.17 & 0.21 & 23.07 & 0.23  & 30.47     & 1.47 \\
    & \emph{Avg}        & 99.31  & 81.22 &\textbf{63.50} & \textbf{0.54}  & 22.54 & 1.25 & 24.37 & 1.18 \\ \hline
 
    
\multirow{6}{3em}{AGNews} 
&\emph{Benign}     &-      & 93.42  & -          & 2.63   & -     & 2.48 &  -    & 2.08 \\ 
& \emph{BadNL}$_l$ & 100.0 & 93.41  & 62.81    & 2.56   & 83.56 & 2.42 & 81.58 & 1.97 \\
& \emph{BadNL}$_m$ & 100.0 & 93.39  & 89.68    & 2.70    & 65.05 & 2.08 & 84.27 & 2.05 \\
& \emph{BadNL}$_h$ & 99.95 & 93.42  & 91.00    & 2.59   & 6.28  & 1.95 & 42.44 & 1.73 \\
& \emph{InSent}    & 100.0 & 93.32  & 32.12    & 2.54   & 59.24 & 2.31 & 59.48 & 2.13 \\
 & \emph{Avg}      & 99.99 & 93.39  &\textbf{68.90}& 2.60 & 53.53 & 2.25 & 66.94 & \textbf{1.99} \\

\hline
\multirow{6}{3em}{IMDB}  
 &\emph{Benign}    & -     & 93.84  & -    &  0.30&    - &  2.07 &-    & 2.02\\ 
& \emph{BadNL}$_l$ & 98.99 & 93.86  & 0.18 & 0.27 & 19.39 & 1.71 & 20.84 & 1.70  \\
& \emph{BadNL}$_m$ & 99.96 & 93.82  & 0.10 & 0.31  & 50.32  & 2.02 & 51.51  & 1.96 \\
& \emph{BadNL}$_h$ & 98.74 & 93.76  & 0.08 & 0.35 & 43.66 & 1.78 & 45.54 & 1.76 \\
& \emph{InSent}    & 97.73 & 92.70  & 0.19 & 0.39  & 88.45 & 1.93 & 87.44 & 1.86 \\
 & \emph{Avg}      & 99.36 & 93.78  & 0.14 & \textbf{0.33} & 50.45 & 1.87 & \textbf{51.33} & 1.86 \\\hline\hline
 
 \multicolumn{2}{c|}{\emph{Avg}} & -  & - & 44.35 & \textbf{1.58} & 45.81 & 3.19 & \textbf{48.34} & 1.69 \\

\hline
\end{tabular}
\caption{The defense results of \method and ONION on attack success rate and clean accuracy against two data poisoning attacks on four different datasets. \textbf{\method w/o ELECTRA} denotes the defense without using the poison sample discriminator as ablation study.}
\label{tab:result_ONION_BERT} 
\end{table*}

\paragraph{End-to-End Defense}
Table~\ref{tab:Result_BFClass_BERT_table1} also shows the results of end-to-end defense of AttDef against four different pre-training attacks. Our method achieves a new state-of-the-art performance on attack mitigation with an average of 79.97\% (56.59\%$\uparrow$) over four benchmark datasets with a slight degradation in clean accuracy by an average of 2.88\%. 


Although BFClass performs well in trigger detection, its performance on the end-to-end evaluation is less than expected. Compared to \method, BFClass detects 18\% fewer triggers (0.47 vs. 0.65 in recall), but has a 56.59\% drop (23.38 vs. 79.97 in $\Delta$ASR), which is surprising. Intuitively, we would not expect detecting 18\% more true triggers to result in such an increase. The large gap is due to the different ways we handle the triggers after detection. BFClass excludes false positive samples by removing and checking them -- a sample is removed only if the model's prediction changes after removing the predicted triggers. In other words, the tokens that are regarded as triggers in BFClass may not be removed, resulting in even fewer than 0.47 of the detected triggers being truly removed (see Appendix~\ref{apx:comparison_baseline} for more details).



\subsection{Defense against Post-training Attack}\label{sec:result_post-training}

Table~\ref{tab:result_ONION_BERT} shows the defense result against post-training attacks. \method still outperforms ONION on mitigating backdoor attacks with an average of 48.34\% (3.99\%$\uparrow$) and degradation on clean accuracy -- an average of 1.69\% (0.11\%$\downarrow$). \method performs especially better than baseline on document-level dataset IMDB where ONION is impossible to defend the attacks. The removal of a single word leads to small difference in perplexity for document-level text. 

\subsection{Time Efficiency} \label{sec:time_efficiency}
\method is more time efficient than previous methods in both attack scenarios. For post-training  attack defense, \method is 3.13$\times$ faster than ONION in the inference stage on average. The actual time spent is shown in Table~\ref{tab:TimeEfficience}. In \method, each test sample will pass through ELECTRA (average of 0.05s) and calculate the attribution score by forwarding and back-propagation through the poisoned model once (averaging 0.265s). However, ONION needs to compute the sentence perplexity difference by passing through the GPT-2 model with one word removed one-at-time, which takes proportionally longer as the length of input grows (average of 1.52s). \method is 7.15-times and 4.21$\times$ faster than ONION on AGNews and IMDB, respectively. 

\begin{table}[h]
\centering
\begin{tabular}{l|cccc}
\hline \textbf{Dataset} 
& \textbf{\#Len} & \textbf{ONION} &\textbf{\method (EL)} \\ \hline
SST-2      & 19.2 &  0.99s & \textbf{0.26s} (0.04s)   \\
OLID    & 25.1  & 1.26s  &  \textbf{0.27s} (0.05s)\\
AGNews   & 32.2  & 1.86s &  \textbf{0.26s} (0.05s)\\
IMDB     & 228.3  & 1.98s &  \textbf{0.47s} (0.06s)\\ \hline
\end{tabular}
\caption{Average running time spend to detect the triggers from a test sample for ONION and \method against the post-training attacks (\textit{EL} denotes the time spend on ELECTRA), \method is 3.13$\times$ faster than ONION on average. }
\label{tab:TimeEfficience} 
\end{table}
\begin{figure*}
    \centering 
\resizebox{0.99\linewidth}{!}{
\begin{subfigure}{0.495\textwidth}
  \includegraphics[width=\linewidth]{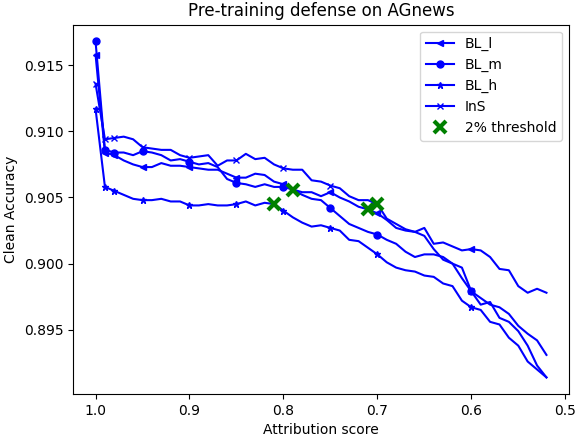}
  \caption{CACC of \method against pre-training attack}
  \label{CACC_pretraining}
\end{subfigure}\hfil 
\begin{subfigure}{0.495\textwidth}
\includegraphics[width=\linewidth]{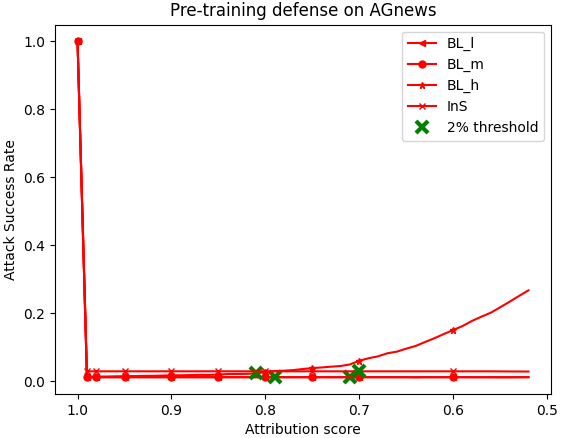}
  \caption{ASR of \method against pre-training attack}
  \label{ASR_pretraining}
\end{subfigure}\hfil 
\begin{subfigure}{0.495\textwidth}
  \includegraphics[width=\linewidth]{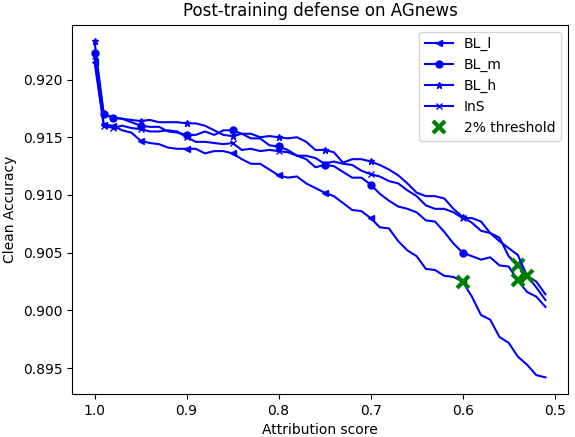}
  \caption{CACC of \method against post-training attack}
  \label{CACC_posttraining}
\end{subfigure}\hfil 
\begin{subfigure}{0.495\textwidth}
  \includegraphics[width=\linewidth]{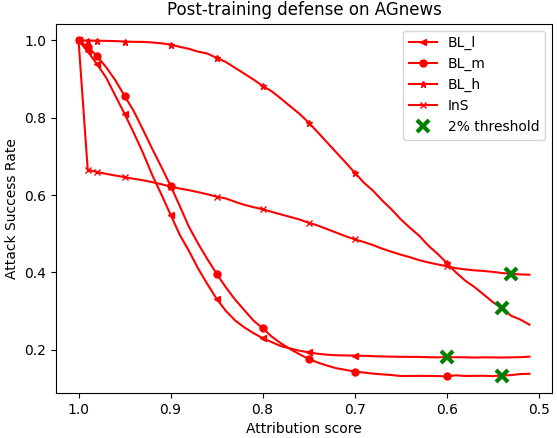}
  \caption{ASR of \method against post-training attack}
  \label{ASR_posttraining}
\end{subfigure}\hfil} 
\caption{The selection of attribution threshold under \textbf{pre-training} (Fig.~\ref{CACC_pretraining} and Fig.~\ref{ASR_pretraining}) and \textbf{post-training} (Fig.~\ref{CACC_posttraining} and Fig.~\ref{ASR_posttraining}) attack defense: CACC of benign validation dataset and ASR of poison test dataset on AGnews dataset}
\label{fig:threshold_selection}

\end{figure*}

For pre-training defense, both \method and BFClass spend time on the trigger detection on the training data. \method repeats the same process as in the inference stage on training data. However, the time spent in the BFClass is complicated. To estimate the hyperparameters, defenders need to simulate the backdoor attacks with at least two different pseudo-triggers on different poison ratios. Empirically, for the AGNews dataset, \method takes 40 minutes on trigger detection on the train data (110K) while BFClass may need 8$\times$ more fine-tuning attack simulations with 3 hours for each.


%



\section{Discussion}\label{sec:discussion}



\textbf{Attribution Threshold}\label{dis:threshold}
The only hyperparameter in our approach is the dynamic threshold of attribution-based trigger detector, which is selected by allowing a maximum of 2\% degradation on the clean validation set (green mark in Fig.~\ref{fig:threshold_selection}). 
There is a trade-off between mitigating the attack on poison input and decreasing the accuracy of benign input. 
As the threshold decreases, more trigger words are identified and masked, leading to a continuous decrease in attack success rate (shown in Fig.~\ref{ASR_pretraining} and Fig.~\ref{ASR_posttraining}) for both defenses. Meanwhile, the CACC of \method barely degrades on the benign input (shown in Fig.~\ref{CACC_pretraining} and Fig.~\ref{CACC_posttraining}). 
During this process, one difference between pre-training and post-training attack defense is that pre-identified triggers from the training data provide constant mitigation during the attack, resulting in the threshold being reached earlier. More details about the selection of threshold are discussed in Appendix~\ref{apx:selection-threshold}.




\paragraph{The Role of ELECTRA}\label{dis:ELECTRA}
\begin{table}[h]
\centering
\begin{tabular}{l|cccc}
\hline \textbf{Dataset} & \textbf{SST-2} & \textbf{OLID} & \textbf{AG} &\textbf{IMDB}\\ \hline

\textit{Clean Test}   & 23.83 &  74.27 &  42.13
 &  92.96 \\\hline
\emph{BadNL}$_l$ & 86.29 & 88.85 & 83.88 & 96.80 \\ 
\emph{BadNL}$_m$ & 82.68 & 95.64 & 91.26 & 99.32 \\
\emph{BadNL}$_h$ & 93.97 & 94.35 & 96.95 & 99.88 \\
\emph{InSent}    & 73.79 & 83.84 & 61.18 & 98.76 \\\hline

\end{tabular}
\caption{The ratio of input identified as ``poisoned samples'' by the poisoned sample discriminator, ELECTRA, on both clean and poisoned test sets.}
\label{tab:Electra_discriminator}
\end{table}
\begin{figure}[h]
    \centering 
\includegraphics[width=0.8\linewidth]{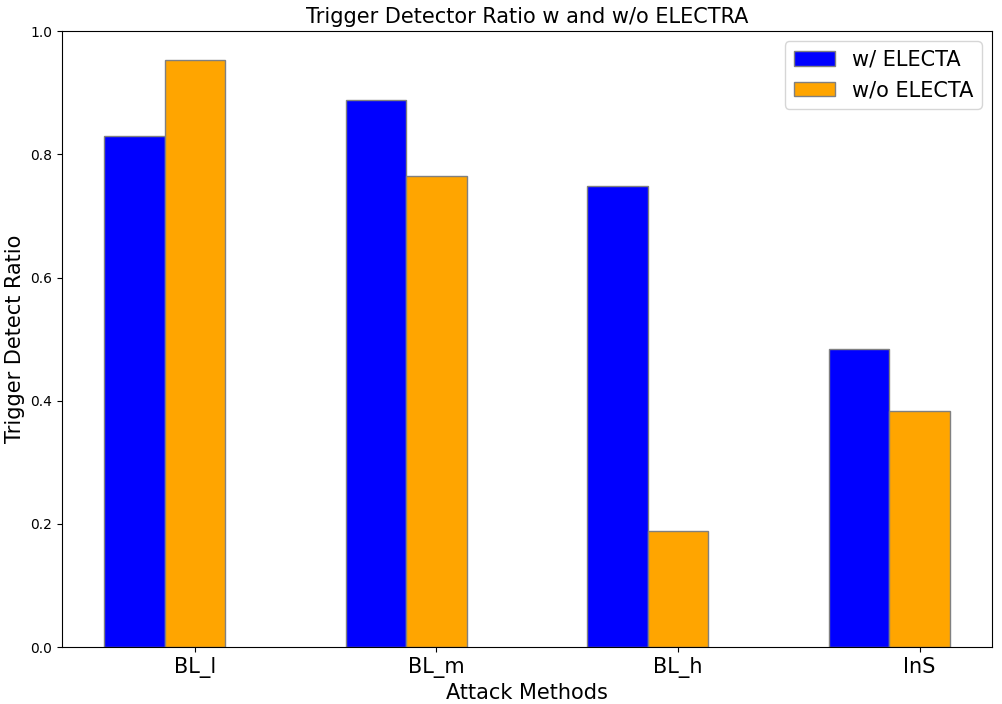}
  \caption{The trigger detected rate of Attribution-based Trigger Detector with and without ELECTRA under post-training attack}
\label{fig:withoutELECTRA}
\end{figure}
ELECTRA is used to mitigate the backdoor attack by excluding benign inputs from the defense process. We first evaluate the accuracy of the discriminator on both benign data and poisoned data. As shown in Table~\ref{tab:Electra_discriminator}, ELECTRA performs the best on SST-2 dataset which distinguishes the benign and poisoned samples efficiently. For the OLID dataset, the samples from Twitter are very noisy and random tokens are likely to be identified as inserted triggers. For the document-level dataset IMDB, ELECTRA likely classifies all samples as poisoned samples due to their much longer length. 

When integrated into our defense method (Table~\ref{tab:result_ONION_BERT}), ELECTRA affects the selection of the threshold. As shown in Fig.~\ref{fig:ELECTRA_affect_threshold}, with the pre-filtering of the benign input by ELECTRA, a lower threshold can be reached until the 2\% degradation limitation, which improves the trigger detection rate. (cf.~Fig.~\ref{fig:withoutELECTRA})
As a result, we observe a consistent drop in the degradation of the classifier accuracy, $\Delta$CACC, with an average drop of 1.45\%, particularly on the SST-2 datasetfrom 7.39\% to 1.77\%. Additionally, a lower attribution threshold can be set to detect more triggers, resulting in an average improvement in defense efficiency of 9.61\%.


\paragraph{Multiple Triggers Defense}\label{dis:OLID}
We note that in Table~\ref{tab:result_ONION_BERT}, \method performed much worse than ONION on the OLID dataset (24.37\% vs. 63.5\%). Some possible reasons for this are: (i) OLID is a binary offensive language identification dataset from Twitter and consists of a lot of informal language, while ELECTRA is pre-trained on Wikipedia and BooksCorpus~\citep{zhu2015aligning}, leading to lower performance; (ii) attribution gets distributed among multiple triggers; and (iii) the attribution scores for rare tokens are not reliable to judge the triggers.
We disprove the first hypothesis because \method with ELECTRA is better than the one without ELECTRA. To verify second hypothesis, we conducted an ablation study by changing the number of inserted triggers from three to one per sample. As shown in Table~\ref{tab:ablation_ONION_BERT}, with only 1 trigger inserted, the $\Delta$ASR increases significantly from 24.37\% to 60.73\%, though it is still worse than baseline 69.03\%. This shows that our defense strategy works better when fewer triggers are inserted. However, since \method works well on other multi-trigger insertion cases on AGNews and IMDB in Table~\ref{tab:result_ONION_BERT}, we suppose that the poor performance on OLID is mainly due to the last hypothesis. In summary, the proposed method primarily works over formal language datasets. Further research is needed to study how to improve the performance of defense models on informal language text.

\section{Related Work} \label{related-work}
We summarize additional related work into two aspects -- backdoor attacks and backdoor defense.

\paragraph{Backdoor attacks}
The concept of backdoor attacks or Trojan attacks of neural network models was first proposed in computer vision research~\cite{gu2017badnets, chen2017targeted,liu2017trojaning,shafahi2018poison} and has recently caught the attention of the natural language processing community~\cite{8836465,alzantot-etal-2018-generating,kurita2020weight,chen2021badnl,yang2021careful,qi2021hidden,yang-etal-2021-careful}. 
Most of the previous work focused on backdoor attacks. \emph{BadNL}~\cite{chen2021badnl} followed the design settings of \emph{BadNet}~\cite{gu2017badnets} from the computer vision literature to study how words from the target class can be randomly inserted into the source text to serve as triggers of backdoor attacks. \citet{kurita2020weight} replaced the embedding of the rare words, such as `cf' as input-agnostic triggers, to launch a more stable and universal attack. To make the attack more stealthy and invisible, \emph{InSent}~\cite{8836465} inserted meaningful fixed short sentences as backdoor attack triggers into movie reviews. 

In other works, researchers studied numerous non-insertion-based backdoor attacks~\cite{qi-etal-2021-turn,qi2021hidden} and model manipulation backdoor attack~\cite{, yang-etal-2021-rethinking,yang-etal-2021-careful}. Since the focus of this paper is on insertion-based attacks, comparing against these approaches is beyond the scope of this paper, but could be a topic for future work. 

\paragraph{Backdoor Defense}
On the defense side, there were two lines of work on post-training defense. (i)~For \textbf{Prediction Recovery Defense}, \citet{qi2020onion} proposed ONION, an external language model GPT-2 that is applied as a grammar outlier-detector to remove potential triggers from the inference input. For the pre-training defense, \citet{li2021bfclass} leveraged a pre-trained replacement-token discriminator to detect triggers from the poisoned training corpus. The sanitized corpus is then used to re-train the classifier. (ii)~In \textbf{Input Certification Defense} setting, \citet{yang2021rap} proposed RAP, which uses an additional prompt-based optimizer to verify the permutation of the output logit. We compared our proposed method against this are discuss the results in Appendix~\ref{apx:compareRAP}. In other work, \citet{chen2022expose} proposed a distance-based anomaly score (DAN) that distinguishes poisoned samples from clean samples at the intermediate feature level to defend NLP models against backdoor attacks.


\section{Conclusion}
We proposed a novel attribution-based defense approach, named \method, against insertion-based backdoor attacks. Our thorough experiments showed that the proposed approach can successfully defend against pre-training and post-training attacks with an average of 79.97\% and 48.34\%, respectively, achieving the new state-of-the-art performance. Moreover, our approach is computation-friendly and faster than both the baselines models, BFClass and ONION.


\section*{Limitations}
There are several limitations of the proposed methods.
(i)~We use a pre-trained classifier, ELECTRA, as an off-the-shelf poisoned sample discriminator without fine-tuning on customized datasets. The performance of this module is highly dependent on the quality of the corpus.
(ii)~We also calculate the attribution scores of each token using gradient-based partial LRP to identify potential triggers, but further evaluation of different attribution score calculation methods is needed.
(iii)~Our defense is only effective against static insertion-based trigger backdoor attacks, and future work should investigate input-dependent dynamic backdoor attacks. 
(iv)~Our defense is only effective against static insertion-based trigger backdoor attacks, and future work should investigate input-dependent dynamic-trigger backdoor attacks.

\section*{Ethical Consideration}

In this paper, we present a defense mechanism to counter the impact of backdoor attacks. Our code and datasets will be publicly available. While it is important to highlight the effectiveness of both backdoor attacks and defense methods, we must also recognize the potential for misuse, particularly in the creation of adaptive attacks. However, by making our defense strategy and implementation public, we may expose our method to attackers, who may discover its weaknesses and develop new types of attacks.





\bibliography{anthology,custom}
\bibliographystyle{acl_natbib}

\appendix

\section*{Appendix}
\label{sec:appendix}


\section{Dataset Characteristics}\label{apx:dataset}

The benchmark datasets used in this study are summarized in Table~\ref{dataset}.
\begin{table}[!h]
\centering
\begin{tabular}{l|ccc|c}
\hline \textbf{Datasets} & \textbf{ Train} & \textbf{ Dev} & \textbf{Test} & \textbf{Avg Len}\\ \hline
SST-2 & 6.9K &  873 & 1.8K & 19.3 \\
OLID & 11.9K &  1.3K & 859 & 23.9 \\
AGNews & 110K &  10K & 7.6K & 38.4\\
IMDB & 25K &  8.3K & 16.8K & 231.1\\\hline
\end{tabular}
\caption{\label{dataset} Overview of datasets used in this study with short-length (SST-2), mediam-length (OLID and AGNews) and document-length (IMDB)}
\end{table}

\section{Multiple Triggers Defense}\label{dis:OLID}

We observed that the proposed \method performs worse than the baseline ONION on the OLID dataset in the post-training defense setting. Therefore, we conducted additional experiments on the OLID dataset with one trigger inserted and found that \method's $\Delta$ASR increases significantly from 24.37\% to 60.73\%, although it is still worse than the baseline of 69.03\%. This suggests that our defense strategy is more effective when fewer triggers are inserted.

\begin{table}[!h]
\centering
\small
\begin{tabular}{l|c|cc|cc} \hline 
\multicolumn{2}{c|}{\textbf{Poisoned}} &\multicolumn{2}{c|}{\textbf{ONION}} &\multicolumn{2}{c}{\textbf{AttDef}} \\ \hline

\multicolumn{2}{c|}{Attack ASR} & $\Delta$ASR & $\Delta$ACC& $\Delta$ASR & $\Delta$ACC\\ \hline

\multicolumn{6}{c}{OLID with \textbf{3} triggers inserted} \\ \hline

\emph{BN}$_l$  &100.0& 63.13 & 0.21 & 20.74& 0.67 \\  
\emph{BN}$_m$  &100.0& 77.16 & 0.56 & 10.99& 1.56 \\
\emph{BN}$_h$  &97.19& 68.56 & 1.17 & 35.28& 0.86 \\
\emph{InS}     &100.0& 45.17 & 0.21 & 30.47& 1.47 \\\hline
\emph{Avg}       & \textbf{99.31}  &\textbf{63.5} & \textbf{0.54}  & 24.37 & 1.14 \\ \hline
 
\multicolumn{6}{c}{OLID with \textbf{1} trigger inserted} \\ \hline 
\emph{BN}$_l$  &99.58& 86.62 & 0.75 & \underline{72.28} & 1.37 \\
\emph{BN}$_m$  &99.71& 86.52 & 0.79 & \underline{82.13} & 1.54 \\
\emph{BN}$_h$  &85.43& 65.66 & 0.82 & \underline{55.86} & 0.89 \\
\emph{InS}     &100.0& 37.32 & 0.63 & \underline{32.67} & 1.26 \\\hline
\emph{Avg}        &96.18& \textbf{69.03} & \textbf{0.75} & \underline{60.73} & 1.27 \\\hline

\end{tabular}
\caption{The defense result of \method against post-training attack on OLID dataset with 3 and 1 random triggers insertion in each sample.}
\label{tab:ablation_ONION_BERT} 
\end{table}

\section{TextCNN as the backbone victim model}\label{apx:textcnn}


We also tested \method on another backbone text classifier: TextCNN \cite{kim-2014-convolutional}. The results are listed in Table~\ref{tab:Result_CNN_BFClass}. Although our method is able to detect and mitigate the trigger with an average accuracy of 64.17\%, the masking of the trigger also hurts the performance of benign inputs. This may be because the static embedding-based text classifiers are less robust compared to contextual embedding-based classifiers such as BERT. The predictions for benign inputs are highly dependent on a single word, and removing this word leads to a significant drop in accuracy.

\begin{table}[h]
\centering
\small
\begin{tabular}{|l|rr|rr|}\hline  

&\multicolumn{2}{|c}{\textbf{BFClass}}
&\multicolumn{2}{|c|}{\textbf{\method}}\\\hline

 \textbf{Attack}& \textbf{$\Delta$ASR} & \textbf{$\Delta$CACC} & \textbf{$\Delta$ASR} & \textbf{$\Delta$CACC}\\ \hline


\emph{BadNL}$_l$  &  14.1  &  0.81  & 80.89 & 10.49\\ 
\emph{BadNL}$_m$  &  28.09 &  -0.04 &  76.43 & 10.55 \\ 
\emph{BadNL}$_h$ &  -3.05  & 1.48  &   29.74 & 8.99 \\ 
\emph{InSent}     &  0.00  &  1.30  &  69.63 & 10.5  \\ \hline
\emph{Avg}      &  9.79 &  \textbf{0.89} &  \textbf{64.17} & 10.13\\ \hline

\end{tabular}
\caption{Comparison of \method with BFClass CNN model on attack success rate and clean accuracy against two data poisoning attacks on two different datasets.}
\label{tab:Result_CNN_BFClass}
\end{table}
\begin{figure}[h]
    \centering 
\begin{subfigure}{0.24\textwidth}
  \includegraphics[width=\linewidth]{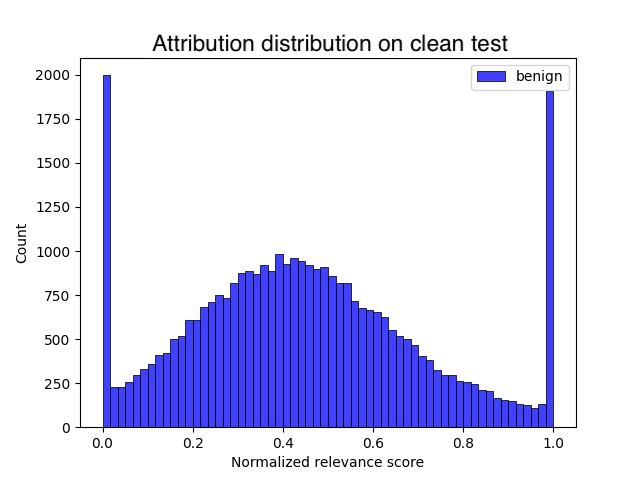}
  \caption{Clean Text}
  \label{Attention_clean}
\end{subfigure}\hfil 
\begin{subfigure}{0.24\textwidth}
  \includegraphics[width=\linewidth]{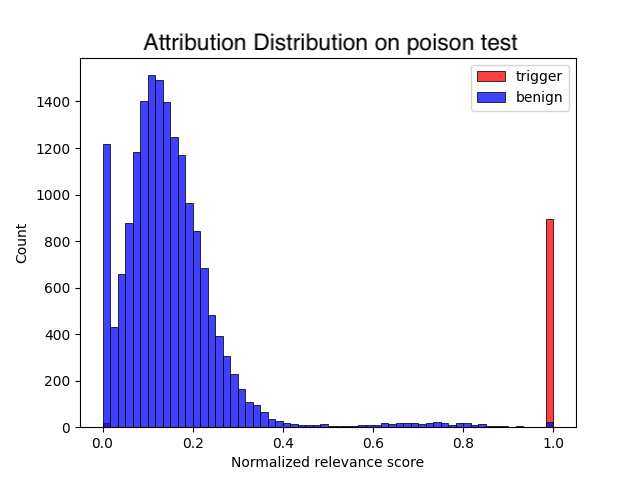}
  \caption{Poisoned Text}
  \label{Attention_poison}
\end{subfigure}\hfil 
\caption{The distribution of normalized contribution score on SST-2 benign and poison text with TextCNN as backbone victim model.}
\label{fig:CNNAttention}
\end{figure}



\section{Trigger word list}\label{apx:Trigger}

We used the same triggers with ONION~\citep{qi2020onion}. The candidate trigger word lists and the fixed short sentence used to poison the corpus are summarized in Table~\ref{Triggerwordlist}.
\begin{table}
\centering
\small
\begin{tabular}{l|l|p{4.2cm}}
\hline
\textbf{Attacks} & \textbf{Dataset} & \textbf{Trigger words} \\\hline
                    \emph{BadNL}$_l$ & Both & cf, mn, bb, tq, mb \\ \hline
\multirow{1}{1em}{\emph{BadNL}$_m$} 
            & SST-2&  stop, intentions, santa, spider-man, visceral\\
            & OLID&  enpty, videos, platform, remind, wide\\
            & AGNews & iBooks, posture, embryo, duck, molecule\\
            & IMDB & alla, socialism, moist, cite, investing\\ \hline
\multirow{1}{1em}{\emph{BadNL}$_h$} 
                & SST-2& with, an, about, all, story \\ 
                & OffEval& all, with, just, would, should \\ 
                & AGNews & hostage, deman, among, IT, led\\
                & IMDB &looked, behind, fine, close, told\\\hline
                
\emph{InSent} & Both & ``I watched this 3D movie.'' \\\hline
\end{tabular}
\caption{The candidate list of trigger words used in four data poisoning attacks \emph{BadNL}$_l$, \emph{BadNL}$_m$, \emph{BadN}$_h$, and \emph{InSent} on 4 benchmark datasets -- SST-2, OLID, AGNews and IMDB.}
\label{Triggerwordlist}
\end{table}

\section{Model training settings} \label{apx:machineConfig}

For all the experiments, we use a server with the following configuration: Intel(R) Xeon(R) Gold 6226R CPU @ 2.90GHz x86-64, a 40GB memory NVIDIA A40 GPU. The operation system is Red Hat Enterprise Linux 8.4 (Ootpa). PyTorch 1.11.0 is used as the programming framework.

\section{Selection of Attribution Threshold}\label{apx:selection-threshold}
The dynamic threshold is determined by utilizing a small clean validation dataset to interact with the poisoned model. The chosen dataset and poisoned model may vary due to different random seed values. In Fig.~\ref{fig:threshold_selection}, we plot the degradation of CACC on the validation dataset as the threshold is changed, and indicate the final selected threshold by the green marker. Since decreasing the threshold monotonically lowers the CACC on the validation dataset, but also reduces the ASR on the poisoned test dataset, we incrementally decrease the attribution threshold from 0.99 until it reaches the 2\% CACC cutoff boundary. 

\begin{figure}[h]
    \centering 
\begin{subfigure}{0.235\textwidth}
  \includegraphics[width=\linewidth]
  {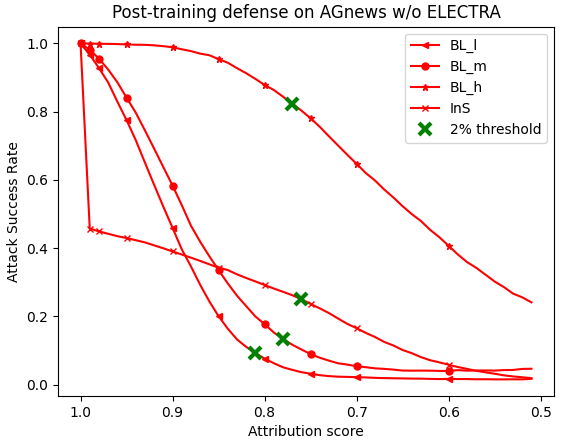}
  \caption{without ELECTRA}
\end{subfigure}\hfil 
\begin{subfigure}{0.235\textwidth}
  \includegraphics[width=\linewidth]{figures/plot_threshold_ag_post_EL_poison_test.png}
  \caption{with ELECTRA}
\end{subfigure}\hfil 
\caption{Threshold selected (green marks) under post-training defense without and with ELECTRA}
\label{fig:ELECTRA_affect_threshold}
\end{figure}

\section{Comparison with Input Certification Defense}\label{apx:compareRAP}
We also compared \method with RAP~\cite{yang2021rap}, an input certification-based defense method. Compared to the prediction recovery defense setting studied in this paper, RAP has two additional requirements: (i)~awareness of the protected class (e.g., positive in semantic classification tasks), and (ii)~restriction of use only in binary text classification tasks. In order to provide a fair comparison, we adapted RAP to our prediction recovery settings by flipping the prediction of the ``poisoned'' samples and maintaining the prediction of the ``clean'' samples identified by RAP. Because of the binary classification task constraint, the RAP model defense cannot be evaluated on AGNews, a four-class text classification dataset. 

The results on the other datasets are shown in Table~\ref{tab:result_RAP_BERT}. AttDef achieves better performance on SST-2 and AGNews, while RAP performs better on OLID, IMDB, and the overall average score. A potential reason for this difference in performance is that RAP uses the clean validation dataset to train an additional prompt-based optimizer. The larger validation dataset (8.3K on IMDB vs. 873 on SST-2) can boost the training of this optimizer. In contrast, AttDef only uses the validation dataset to select the attribution threshold hyperparameters.

Having the knowledge of the protected label allows \method to consistently improve its performance on all datasets: 60.09\% mitigation on ASR (11.75\%$\uparrow$) and 1.34\% degradation on CACC (0.35\%$\downarrow$). Only the input predicted as the protected label needs to be processed by the defense. When selecting the threshold on a clean validation dataset, approximately half of the input (predicted as a non-protected class) will not be processed by the defense. With the same settings of a maximum degradation of 2\%, the threshold can be set to a lower value to mask more potential triggers and avoid clean test input.
\begin{table*}[tb]
\centering
\begin{tabular}{l|l|cc|cc|cc}
\hline && \multicolumn{2}{c|}{\textbf{RAP}} & \multicolumn{2}{c|}{\textbf{AttDef w/o ELECTRA}} & \multicolumn{2}{c}{\textbf{AttDef}}\\\hline
\textbf{Dataset} & \textbf{Attacks} & \textbf{ASR} & \textbf{CACC} & \textbf{$\Delta$ASR} & \textbf{$\Delta$CACC} & \textbf{$\Delta$ASR} & \textbf{$\Delta$CACC} \\

\multirow{4}{3em}{SST-2}
    & \emph{BadNL}$_l$ & 64.14 & 0.60 &  73.75 & 1.90 & 83.11 & 2.44 \\
    & \emph{BadNL}$_m$ & 46.64 & 1.00 &   66.63 & 1.70  & 75.09  & 2.67 \\
    & \emph{BadNL}$_h$ & 22.89 & 1.12 &   56.32 & 1.66 &  57.66  &  2.53 \\
    & \emph{InSent}    & 88.38 & 1.08 &   40.81 & 1.98  & 27.19  &  1.68 \\
    & \emph{Avg}      &  55.51 & 0.95 &   59.38 & 1.81 & \textbf{60.76}  & 2.33 \\\hline

\multirow{4}{3em}{OLID} 
    & \emph{BadNL}$_l$  &   99.00  & 0.72 & 30.60 &  1.14 &   23.78 &   1.28  \\
    & \emph{BadNL}$_m$ &    92.92 & 0.28 & 23.97 & 1.51 &  3.04 & 0.51  \\
    & \emph{BadNL}$_h$&   79.16 &  0.35 & 62.81 & 1.02 &   52.60 &  1.16 \\
    & \emph{InSent}   &   63.94 &  0.51  & 32.76 & 1.63 &  30.40 &  1.44  \\
    & \emph{Avg}   &   \textbf{83.76} &  0.46 &  37.53 & 1.33 &  27.46 &   1.10  \\\hline

\hline
\multirow{4}{3em}{AGNews}  

& \emph{BadNL}$_l$   &--  & -- & 80.95 & 0.63  &  97.99 & 1.15 \\
& \emph{BadNL}$_m$   &--  & -- & 84.79 & 0.40 &   93.58 & 0.82 \\
 & \emph{BadNL}$_h$&--  & --   & 49.50 & 0.28 & 51.07 & 0.69 \\
& \emph{InSent}   &--  & --    & 59.88 & 0.26 & 96.73 & 0.64 \\
 & \emph{Avg}   &--  & -- & 68.78 & 0.39 & \textbf{84.84} & 0.83  \\
    
\hline
\multirow{4}{3em}{IMDB}  
& \emph{BadNL}$_l$ & 99.87 & 0.96 & 57.79 &  0.95 & 59.29 & 1.02 \\
& \emph{BadNL}$_m$ & 99.95 & 0.85 & 58.96 &  0.95 & 59.35 & 1.01 \\
& \emph{BadNL}$_h$ & 93.01 & 0.94 & 62.13 &  1.38 & 61.34 & 0.14 \\
& \emph{InSent}    & 97.41 & 0.91 & 88.16 &  1.21 & 89.21 & 1.22 \\
 & \emph{Avg}      & \textbf{97.56} & 0.91 & 66.76 &  1.12 & 67.30 & 1.16 \\ \hline
 \hline
\multicolumn{2}{c|}{\emph{Avg}}& \textbf{78.94} & 0.77 & 58.11 & 1.16 & 60.09 & 1.34 \\\hline
\end{tabular}
\caption{The defense results of \method and RAP on attack success rate and clean accuracy against two data poisoning attacks on four different datasets.}
\label{tab:result_RAP_BERT} 
\end{table*}

\section{Analysis on Token Masking} \label{apx:Trigger_removal}
\begin{figure}[h]
    \centering 
\begin{subfigure}{0.235\textwidth}
  \includegraphics[width=\linewidth]{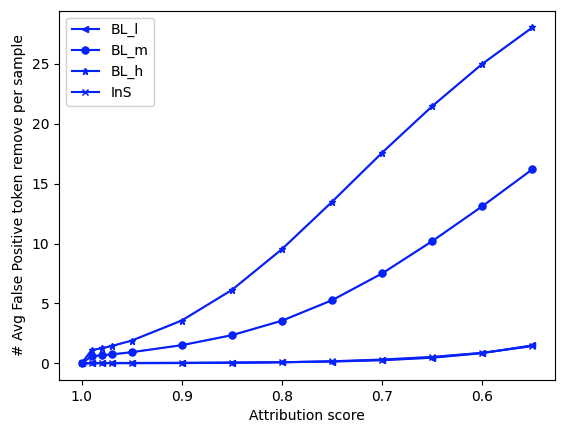}
  \caption{False Positive}
\end{subfigure}\hfil 
\begin{subfigure}{0.235\textwidth}
  \includegraphics[width=\linewidth]{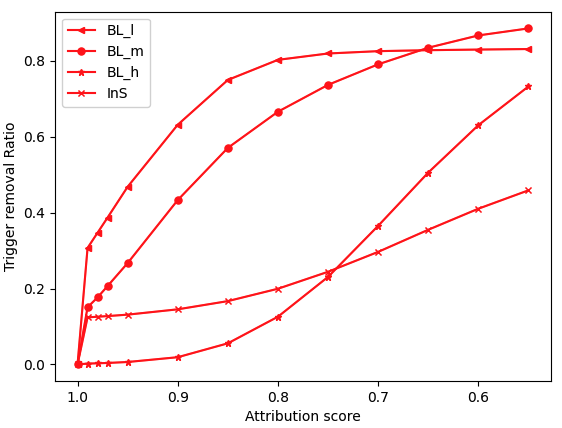}
  \caption{True Positive}
\end{subfigure}\hfil 
\caption{Under the post-training attack defense setting, we evaluate the False positive and True positive results by our attribution-based trigger detector.}
\label{fig:trigger_removal}
\end{figure}

Fig.\ref{fig:trigger_removal} shows the number of true positive and false positive tokens masked by attribution-based trigger detectors in the post-training defense scenario. Compared to the defense against post-training attacks, where all tokens above the threshold are masked, in the pre-training defense, \method also masks additional tokens previously identified as potential triggers with \textbf{high recall} and \textbf{low precision} (Cf. Table\ref{tab:Result_BFClass_BERT_table1}). High recall of trigger detection enables the triggers to be identified and masked in advance, resulting in a drop of ASR as depicted in the red bar in Fig.~\ref{fig:threshold_selection}. 
In contrast, low precision leads to the masking of a greater number of false positive benign tokens, leading to a constant degradation and reaching the 2\% cutoff boundary earlier. Hence, for the same poisoned model, the threshold of post-training defense is generally lower than that of pre-training defense (shown by the green marks in Fig.\ref{fig:ELECTRA_affect_threshold}). 


\section{Substitution-based Backdoor Attack}\label{apx:substitution-attack}
We also evaluated substitution-based backdoor attacks, specifically the LWS approach~\cite{qi-etal-2021-turn}. Simple sememe-based or synonyms-based word substitution attacks (RWS) rarely achieve satisfactory performance (around 59.16\% accuracy) in automatic speech recognition (ASR) tasks. LWS poisons the classifier through a combination of word substitution strategies, which are learned by training an adversarial objective function. Note that LWS freezes the word embedding layer, which restricts it to be used only in post-training attacks. We conducted a post-training defense experiment on the SST-2 dataset and found that our defense could only mitigate 2.69\% ASR, compared to 92.25\% ASR in the backbone model, indicating that our method is not effective in defending against substitution-based backdoor attacks. Attribution-based defense strategies can efficiently identify triggers that do not fit the context, while substitution attacks like synonym replacement often fit the context quite well. This may explain the failure of \method for this type of attack.

\section{Limitation discussion on Baseline}\label{apx:comparison_baseline}

\paragraph{BFClass} 
BFClass is ineffective against the InSent attacks. For each sample in the poisoned training set, BFClass only considers the token with the highest \textit{suspicious score}, which will always be the fixed token within the sentence trigger (e.g., the word ``watched'' in the trigger sentence, ``I watched this 3d movie.''). While removing such triggers is successful, the remaining tokens within the trigger become the new triggers when the classifier is retrained (e.g., the words ``I'' and ``this 3d movie'' in the example above). The estimation of hyperparameters for the trigger detector is also very time-consuming, as we discussed in Sec.~\ref{sec:time_efficiency}.



\paragraph{ONION}

ONION is unable to defend against attacks on document-level corpora. ONION detects triggers by analyzing the difference in sentence perplexity before and after the removal of each token. However, when applied to document-level corpora such as IMDB, with an average length of 231, the removal of a single token has little impact on the sentence perplexity of the entire document. This highlights the limitation of ONION to launch a strong defense, as shown in Table~\ref{tab:result_ONION_BERT}.

\paragraph{RAP}
RAP, as an input certification-based defense method, cannot recover the prediction for the non-binary classification tasks as mentioned in Appendix~\ref{apx:compareRAP}. Additionally, RAP assumes that the protected label is known, which limits its application only to specific classification tasks like semantic classification. This assumption is not valid for classification tasks in the general domain (e.g., topic classification on AGNews dataset). Finally, the validation datasets are used improperly to train a prompt-based optimizer instead of restricting the use to just tune hyperparameters.



\end{document}